\documentclass[graybox]{svmult}

\usepackage{type1cm}        

\usepackage{makeidx}         
\usepackage{graphicx}

\usepackage{multicol}        
\usepackage[bottom]{footmisc}

\usepackage{newtxtext}       %
\usepackage{newtxmath}       

\makeindex

\usepackage{hyperref}
\usepackage{physics}
\usepackage{natbib}
\usepackage{mathtools}
\usepackage{dsfont}
\usepackage{float}
\usepackage{multicol}
\usepackage{algorithm2e}
\usepackage{subfig}
\usepackage{todonotes}

\usepackage[switch]{lineno}

\newcommand{\X}{\mathbf{X}}
\newcommand{\Z}{\mathbf{Z}}

\title*{Application of the Singular Spectrum Analysis on
  electroluminescence images of thin-film photovoltaic modules}
\titlerunning{Application of SSA on EL images of thin-film PV}

\author{Evgenii Sovetkin and Bart E. Pieters}
\authorrunning{E. Sovetkin and B.E. Pieters}

\institute{Evgenii Sovetkin and Bart E. Pieters
  \at IEK5-Photovoltaik, Forschungszentrum J\"ulich,
  52425   Germany \\
  \email{e.sovetkin@fz-juelich.de}}

\begin{document}
\linenumbers
\let\makeLineNumber\relax

\maketitle

\abstract{This paper discusses an application of the singular spectrum
  analysis method (SSA) in the context of electroluminescence (EL)
  images of thin-film photovoltaic (PV) modules. We propose an EL
  image decomposition as a sum of three components: global intensity,
  cell, and aperiodic components. A parametric model of the extracted
  signal is used to perform several image processing tasks. The cell
  component is used to identify interconnection lines between PV cells
  at a sub-pixel accuracy, as well as to correct incorrect stitching
  of EL images. Furthermore, an explicit expression of the cell
  component signal is used to estimate the inverse characteristic
  length, a physical parameter related to the resistances in a PV
  module.}


\vspace{24pt}

\section{Introduction}

There has been an increasing interest in automated image analysis of
spatially resolved characterisation methods for photovoltaic (PV)
modules such as electroluminescence (EL) \cite{demant2018deep,
  deitsch2018segmentation, deitsch2019automatic, de2019automatic,
  karimi2019automated, sovetkin2019automatic, sovetkin2020encoder,
  sovetkin2020pvaided}. Such automated image analysis aims at quality
control of modules and is thus of great interest for manufacturers, PV
system owners, and insurance companies, as it allows for a systematic
inspection of a large number of modules, both prior and after
installation.

Electroluminescence is a commonly used imaging technique for PV
modules. It relies on the reciprocal operation of the photovoltaic
module as a light-emitting diode, so instead of generating an electric
current from light, an electric current is driven through the solar
cell, which then emits light. As generating electricity and emitting
light are reciprocal processes, one process reveals much about the
other. Electroluminescence (EL) images provide spatially resolved
information on the solar module and is commonly used to locate and
identify defects in the device or extract other (local) solar cell
properties.

In this paper our focus lies on EL images of thin-film PV modules. For
thin-film technology, unlike for a more common crystalline silicon,
little is known about shapes and appearances of defects in EL
images. For the crystalline silicon PV modules there exists a
well-established catalogue of defects visible in EL images (see
\cite{kontges2015iea}), whereas such a catalogue does not exist for
thin-film modules.

To study defects in EL images, it is important to find a compact way
to represent EL image data. This paper proposes such an approach and
considers several image processing algorithms for EL images of
thin-film PV modules that are based on the singular spectrum analysis
(SSA). Our contributions here are manifold.

Firstly, a specific grouping in the SSA algorithm decomposes an EL
image into several components: global intensity variation component,
local periodic intensity component (or cell component), and a residual
image that contains various local aperiodic features. We argue that
each of these components has a different physical origin.

Secondly, the extracted components of an EL image can be approximated
by a parametric model, that represents an EL image as a small
dimensional vector, and hence our methods can also be considered as a
dimensionality reduction technique.  Furthermore, the parametric model
of the cell components is used to estimate the position of the
interconnection line between individual PV cells. Our algorithm
features symbolic differentiation and estimates positions of the
interconnection lines at a sub-pixel accuracy. A similar technique is
used in the estimation of the inverse characteristic length, a
physical characteristic of a PV module that equals the square of the
ratio between different resistances in a module.

Lastly, the cell component signal is used to estimate a non-linear
transformation of an image to adjust an incorrectly stitched image.

The rest of the paper is organised as
follows. Section~\ref{sec:methods} reviews the methods and the
corresponding literature used in this paper.  Section~\ref{sec:data}
describes the data used in the project. The main contribution of this
paper is given in Section~\ref{sec:applications}, which focuses on
various applications of SSA to the EL images of thin-film modules.
Lastly, the paper is concluded in Section~\ref{sec:conclusions}.

\section{Methods overview}
\label{sec:methods}

This section describes the methods used in this paper and overviews
related literature. Our main tool is the singular spectrum analysis
method (SSA).

The history of SSA can be traced to the works of
\cite{broomhead1986extracting}, where an SSA-like method was
established and applied in the context of nonlinear dynamics for the
purpose of reconstructing the attractor\linelabel{dynamical system
  analysis and SSA} of a system from measured time series. Further, in
the context nonlinear dynamical system, SSA can be also used for phase
space reconstruction algorithm, \cite{Elsner1996}.

The so-called ``Caterpillar'' methodology is a parallel development of
SSA that originated in the former Soviet Union, especially in Saint
Petersburg, independently of the mainstream SSA work in the West,
\cite{danilov1997principal}. This methodology became known to the rest
of the world more recently.  ``Caterpillar-SSA''
\cite{golyandina2001analysis}, emphasises the concept of separability,
a concept that leads, for example, to specific recommendations
concerning the choice of the SSA parameters.

Originally, the SSA method was applied to the one-dimensional
time-series data. In fact, by now, it is not easy to find an applied
area related to the analysis of temporal data, where one-dimensional
SSA is not being applied. To name a few applications, the method found
its way to the analysis of climate and atmospheric data,
\cite{vautard1989singular,fraedrich1986estimating}, to meteorological
data, \cite{weare1982examples}, as well as to the marine science,
\cite{colebrook1982continuous}. \cite{lima2016gap} used SSA for gap
filling in precipitation data. This method has been also applied in
the financial sector to discover hidden economic cycles,
\cite{sella2016economic}. \cite{groth2015monte} used a multivariate
extension of SSA and defined a Procrustes test to the analysis of
interannual variability in the North Atlantic sea surface
temperature. For further references to various applications of SSA in
time-series analysis see \cite{zhigljavsky2010singular,
  golyandina2018singular}.


More recently, SSA was also used to analyse digital images and other
objects that are not necessarily of planar or rectangular form. This
particular development is utilised in this paper. 
\cite{rodriguez2010singular}, used SSA to define a distance between
images with a possible application in face verification. In
\cite{zabalza2014singular,zabalza2015novel,qiao2016effective} the
1D-/2D-SSA variants were used in the context of hyperspectral images
for the purpose of denoising, feature extraction, and classification
tasks. In \cite{mamou2007singular} SSA was applied in the context of
ultrasonic imaging for improving the imaging of brachytherapy seed. In
application related to geoscientific data, 2D-SSA was utilised for
gap-filling, \cite{zscheischler2014extended}. 2D-SSA was also applied
in texture classification \cite{monadjemi2004towards}, seismology
\cite{trickett2008f}, gene expression \cite{holloway2011gene}, and
medical imaging \cite{shin2014calibrationless}.

In our application of the 2D-SSA, we utilise the ability of the method
to separate signal into trend and periodic components. Further, we use
the parametric form of the extracted signal to perform various image
processing tasks.

In order to explain our algorithms, we give a review of necessary
theory in the following subsections.  The SSA algorithm is reviewed in
Section~\ref{sec:ssa}. SSA itself is non-parametric, however, a
parametric model can be given to describe the extracted signal. There
are two sets of parameters to be estimated in the model. The ESPRIT
method (Section~\ref{sec:esprit}) estimates frequencies and damping
factors of a signal, where the least squares provides an estimation of
the amplitude and phases (Section~\ref{sec:ampl-phase-estim}). Lastly,
remarks on implementation and comparison to similar methods are
discussed in Section~\ref{sec:impl-comp-altern}.

\subsection{SSA}
\label{sec:ssa}

Singular Spectrum Analysis (SSA) is a model-free time series analysis
method that belongs to the so-called subspace methods,
\cite{van1993subspace}. In subspace methods, a signal estimation is
performed by taking a certain linear subspace. SSA can also be
considered as a low-rank approximation method,
\cite{markovsky2018low}.

The general theory of SSA for one-dimensional time series is
elaborated in
\cite{golyandina2001analysis}. \cite{golyandina2018singular} provide
more updated information on extensions with a strong focus on
applications.


The output of SSA-like methods is a decomposition of an observed
signal $x$ (e.g.\ a time series, a multivariate time series or an
image) into a sum of identifiable components:
\begin{equation}
  \label{eq:5}
  x = x_1 + \ldots + x_n.
\end{equation}

Among all SSA-like methods, the following four common algorithm steps
can be isolated.
\begin{enumerate}
\item {\bf Embedding.} The original signal $x$ (e.g.\ time series or
  an image) is mapped into a matrix $\X$, that is called a {\it
    trajectory matrix}.

  The embedding is parametrised by a single parameter denoted by $L$
  (a number or a vector).
\item {\bf Decomposition.}  The second step consists of the
  decomposition of the trajectory matrix $\X$ into a sum of matrices
  of rank 1.

  Often the singular value decomposition (SVD) is used for this
  purpose, which is an optimal rank-one matrix decomposition in the
  Frobenius norm sense.

\item {\bf Grouping.} The third step is a grouping of the
  decomposition components. At the grouping step, the elementary
  rank-one matrices are grouped and summed within groups.

  The grouping algorithm step is often semi-automatic and depends on
  the type of data and application.
\item {\bf Reconstruction.} The grouped components are not necessarily
  valid trajectory matrices. Hence a projector operator to trajectory
  matrix space is applied on the grouped components, resulting in the
  final signal decomposition~(\ref{eq:5}).
\end{enumerate}

In this paper, we utilise the 2D-SSA variant of the algorithm,
\cite{golyandina2009algebraic,golyandina20102d,golyandina2015multivariate}. For
this variant, the trajectory matrix is a Hankel-block-Hankel matrix
and the corresponding embedding is parametrised by a two-dimensional
parameter $L = (L_{1},L_{2}) \in \mathbb{N}^{2}$. The decomposition is
performed with SVD, and the reconstruction projection operator is a
2-step diagonal averaging procedure.  The precise forms of the
embedding, SVD, and reconstruction projection operator are provided in
the Appendix.

The grouping step of the SSA algorithm determines the form of the
final signal decomposition.  A typical SSA decomposition of a signal
is the decomposition into a slowly-varying trend, regular
oscillations, and noise. An important notion in the SSA theory is the
notion of {\it the signal of finite rank}. This notion allows us to
classify and group together rank-one matrices into a trend,
oscillations, and noise components.

Informally, finite rank signals are those that have a trajectory
matrix of a finite rank (see a formal definition in the
Appendix). 2D-SSA produces a class of signals of specific objects of
finite rank. Those objects have the form described in the following
theorem, \cite{golyandina2009algebraic,golyandina2015multivariate}.

\begin{theorem}
  \label{th: finite rank series}
  And infinite 2D-array $\{x(n,m)\}_{n,m \in \mathbb{N}}$ is of finite
  rank if and only if
  \begin{equation}
    \label{eq:2}
    x(n,m) = \sum_{k=1}^K p_k(n,m)
    \rho_{1k}^{m} \rho_{2k}^{n}\cos\big(2 \pi (\omega_{1k}n +
    \omega_{2k}m) + \phi_{k}\big), \quad n,m \in \mathbb{N},
  \end{equation}
  where $p_{k}(n,m)$ are polynomials in $n$ and $m$ variables,
  $\rho_{\cdot k}$ are the damping factors, $\omega_{\cdot k}$ the
  frequency parameters and $\phi_{k}$ are the phase parameters.
\end{theorem}


\subsection{ESPRIT}
\label{sec:esprit}

Estimation of Signal Parameters via Rotational Invariance Techniques
(ESPRIT) is a method to estimate parameters of a mixture of
one-dimensional, \cite{roy1989esprit}, and two-dimensional
amplitude-modulated sinusoids, \cite{rouquette2001estimation,
  wang2005comments}, in background noise.

For the 2D-ESPRIT method, \cite{rouquette2001estimation}, the observed
data $y$ is generated by the following additive model:
\begin{equation}
  y(m,n) = x(m,n) + \varepsilon(m,n),
\end{equation}
where $0\leq m \leq N_{x} - 1$ and $0\leq n \leq N_{y}-1$,
$\varepsilon$ is the zero-mean Gaussian noise with variance
$\sigma^{2}$. The model for the signal $x$ is given by the sum of
amplitude modulated two-dimensional sinusoid
\begin{equation}
  \label{eq:1}
  x(m,n) = \sum_{k=1}^{K} s_{k} \rho_{1k}^m \rho_{2k}^n
  \cos\big( 2 \pi (\omega_{1k}m + \omega_{2k}n) + \phi_{k} \big),
\end{equation}
where $\omega_{1k}, \omega_{2k}$ are the normalised frequencies in
different directions, $\alpha_{1k}, \alpha_{2k}$ are the damping
factors, $s_{k}$ amplitudes, and $\phi_{k}$ phases.

By Theorem~\ref{th: finite rank series} the signal (\ref{eq:1}) is of
finite rank. The ESPRIT methods utilises the fact of the
rank-deficiency of the trajectory matrix of the observed signal $y$,
and a certain transformation matrix between sub-trajectory matrices is
obtained. The frequency and damping factor parameters are computed
from the argument and absolute values of the complex-valued
eigenvalues of the obtained transformation matrix. See more details in
\cite{roy1989esprit,rouquette2001estimation}.


\subsection{Amplitude, phase estimation}
\label{sec:ampl-phase-estim}

The ESPRIT estimates the frequencies $\omega_{\cdot k}$ and the
damping factors $\rho_{\cdot k}$ of the signal model
(\ref{eq:1}). Here only the amplitudes $s_{k}$ and phases $\phi_{k}$
remain to be estimated. This problem can be reformulated as a linear
regression model using the formula of the cosine of sums. With this
formula, (\ref{eq:1}) can be rewritten as
\begin{eqnarray}
  x(m,n) = \sum_{k=1}^{K} A_k \rho_{1k}^{m} \rho_{2k}^{n} \cos(2 \pi
  (\omega_{1k}n + \omega_{2k}m)) \nonumber\\
  - B_k \rho_{1k}^{m} \rho_{2k}^{n}
  \sin(2 \pi (\omega_{1k}n + \omega_{2k}m)),
\end{eqnarray}
where $\rho_{\cdot k}, \omega_{\cdot k}$ are the parameters estimated
from the ESPRIT, and $A_{k}, B_{k}$ are the parameters of the linear
model to be estimated.

Note that the dependent variable in the linear regression model are
the values of $x(n,m)$, the finite-rank signal extracted from $y$. In
terms of the SSA algorithm this corresponds to the series
reconstructed from the selected components.

The amplitude and phase are given by:
\begin{equation}
  s_k = \sqrt{A_k^2 + B_k^2}, \quad \phi_k = \atan(B_k/A_k).
\end{equation}

\subsection{Implementation. Comparison to alternatives methods}
\label{sec:impl-comp-altern}

For our application, we use an R-package ``Rssa'',
\cite{golyandina2018singular,korobeynikov2010computation,golyandina2014basic,golyandina2015multivariate},
where all the required functionality including the ESPRIT method is
implemented. It should be noted that the trajectory matrix for 2D-SSA
is large and has $O(N^{2})$ number of elements, where $N$ is a number
of pixels in the signal image. However, the structure of the
Hankel-block-Hankel matrix allows implementing of SVD with Lanczos
algorithm efficiently in time and memory by computing product of a
matrix and a vector with Fast Fourier Transform,
\cite{korobeynikov2010computation,lanczos1950iteration}.

For computing amplitude and phase parameters, we utilise a standard
R-function ``lm'', \cite{chambers1992linear}.

The form of the parametric model of the SSA signal suggests that
Fourier analysis can be used to obtain similar results. However, in
order to obtain a compact representation of a signal a small Fourier
coefficient should be discarded. For that purpose a sparse Fourier
analysis approach can be used, \cite{hassanieh2012simple}.

However, the parametric model of the SSA signal is more flexible, as
every periodic has an amplitude modulation. Furthermore, the Fourier
analysis is a low-resolution type method, as in the context of time
series, frequency can be estimated only up to $1/N$, where $N$ is the
length of the time series. Whereas ESPRIT is regarded as a
high-resolution method.

\section{Data}
\label{sec:data}

In this section we discuss the data that is used in this paper. The
data was acquired within the framework of the PEARL-TF project. The
\cite{pearltf_eu} website contains detailed information about the
project and the involved partners. In this project, the data from
several solar parks with thin-film modules were collected. In addition
to EL images, also performance characteristics of the modules were
measured.

The EL images are taken at predefined conditions (selected fixed
applied current and/or fixed applied voltage). A silicon CCD sensor
camera is used to measure subsequently several parts of the module,
with the images being stitched afterwards. The applied voltage and the
applied current together with the temperature of the module are being
recorded.  The I/V characteristics are also measured and the solar
cell performance parameters are determined.

\begin{figure}
  \centering
  \includegraphics[angle=90,width=0.9\linewidth]{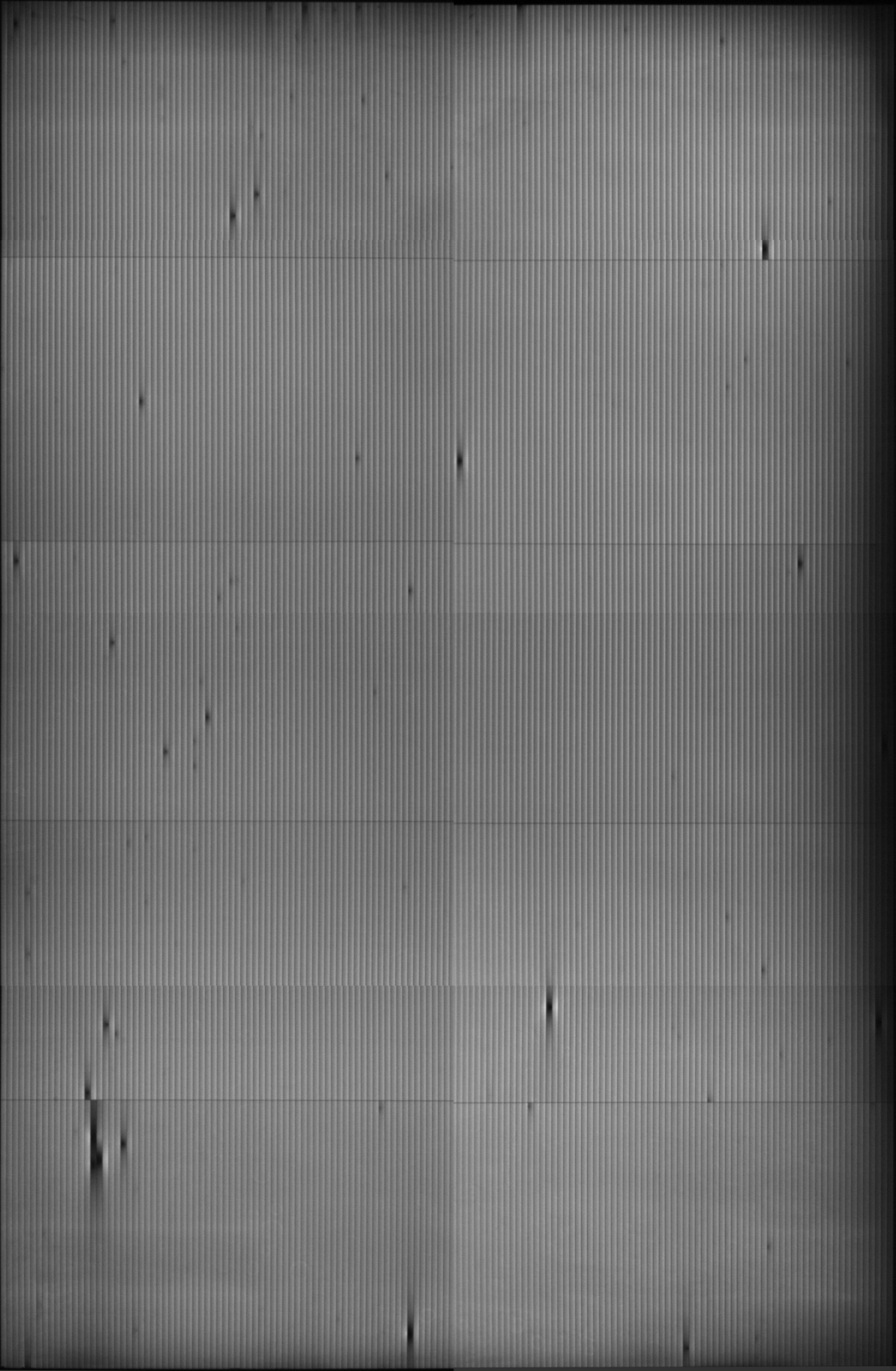}
  \caption{Thin-film module EL image. A module consists of 150 cells
    (positioned horizontally) connected in series. The cells are
    separated by interconnection lines (horizontal dark lines). The
    module consists of several submodules separated by vertical
    isolation lines, which appear dark in the EL image. The EL image
    is stitched (there are 1 horizontal and 3 vertical stitch lines);
    overall intensities of different patches of images are different.
    These intensity differences are attributed to metastable changes
    during the measurement.}
  \label{fig:module}
\end{figure}

The database contains over 9500 EL images of thin-film PV modules.
The bulk of these EL images (about 6000) are from co-evaporated Copper
Indium Gallium di-Selenide (CIGS) modules with a chemical bath Cadmium
Sulfide buffer. All EL images shown in this work are from such CIGS
modules.  Every image is supplied with measured performance data. A
typical EL image of a thin-film module from our database is depicted
in Figure~\ref{fig:module}. The module consists of 150 connected cells
in series (in Figure~\ref{fig:module} the cells are recognised as
horizontal stripes). The cells are separated by interconnection lines
(horizontal dark lines in Figure~\ref{fig:module}). In addition, the
module is separated into 5 parallel sub-modules by vertical isolation
lines (dark vertical lines).

Every EL image consists of several stitched images.  Different
stitched parts of the image have different overall intensities (see
Figure~\ref{fig:module}). This is attributed to the metastable
behaviour of thin-film solar cells, where the electrical properties of
the cell can change during the measurement.

\section{Results}
\label{sec:applications}

In this section we discuss a selection of image processing methods
that we build on top of the SSA framework.

Firstly, Section~\ref{sec:image-decomposition} describes a
decomposition of EL images into several components: global intensity,
cell, and aperiodic components. This is a direct result of the SSA
decomposition, where grouping is performed using prior knowledge of
the image size and number of cells in a module.

Secondly, a parametric model of the cell component signal is used to
achieve several goals. The symbolic differentiation allows a global
search for the minimum points in the cell components, that can be used
to identify the interconnection lines
(Section~\ref{sec:laser-line-detection}). Furthermore, we demonstrate
the estimation of the inverse characteristic length, a physical
parameter that depends on several resistances in a PV module
(Section~\ref{sec:char-length-estim}).

Lastly, MSSA and its parametric model is used to obtain a non-linear
transformation needed to correct the stitch line in EL images.

\subsection{Image decomposition}
\label{sec:image-decomposition}

The grouping step of the SSA algorithm allows combining of decomposed
rank-one components into groups. We define these groups that result in
EL image decomposition onto 3 components: global intensity variation,
cell, and aperiodic components.

\begin{figure}
  \sidecaption
  \centering
  \includegraphics[width=.3\textwidth]{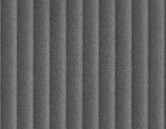}
  \caption{An enlargement of a 10-cells EL image region. Each cell
    exhibit periodic behaviour.}
  \label{fig: blowup cell}
\end{figure}

Figure~\ref{fig: blowup cell} shows a close up image of the module,
that clearly identifies a set of 10 parallel cells separated by
interconnection lines. The EL intensity varies systematically over the
cell width. This is the result of the series resistance of the
electrodes. As the transparent zinc-oxide front electrode exhibits a
much larger sheet resistance than the molybdenum back contact, the
voltage over the diode junction drops from one side to the other
\cite{helbig2010quantitative,pieters2015cellintensity}. As the EL
intensity depends primarily on the cell voltage this leads to a clear
intensity gradient over each\linelabel{physical reason for periodic}
cell.

To achieve good separability of the periodic components, the parameter
$L$ of the SSA embedding is selected to equal approximately half of
the dimensions of the input image,
\cite{golyandina2001analysis,golyandina2018singular}. The ESPRIT
estimates frequencies and damping factors of the components, which are
used in the decision of the grouping step.

A thin-film module consists of a fixed number of interconnected cells,
therefore, the cell components can be identified as periodic
components that have a period smaller than
$\frac{\text{image width}}{150}$. The other components are grouped
together as the global variation and the residual image captures
information of non-low rank signals, such as aperiodic features of an
EL image.

The image decomposition algorithm steps are described in the
Algorithm~\ref{algo:decomposition}.

\begin{algorithm}
  {\bf Input:} EL image $X$ with dimensions $N_{x} \times N_{y}$.

  {\bf Output:} Three images with the same dimensions as $X$: global
  intensity component $G$, cell component $S$ and aperiodic component
  $R$.

  \begin{enumerate}
  \item Perform embedding and decomposition steps of the 2D-SSA
    algorithm for $X$. Compute first $50$ elements in the rank-one
    decomposition of SVD, denote the sum of this components as
    $\tilde \X$.
  \item Apply ESPRIT on the low-rank signal $\tilde \X$.
  \item
    \begin{itemize}
    \item Set the cell components $S$ as in (\ref{eq:1}), choosing
      only periodics with frequency $\omega_{1k} > \frac{150}{N_{x}}$.
    \item The global intensity component $G$ is composed from
      components of non-periodics and periodics with frequency
      $\omega_{1k} < \frac{150}{N_{x}}$.
    \item The $R \coloneqq X - S - G$.
    \end{itemize}
  \end{enumerate}
  \caption{EL image decomposition\label{algo:decomposition}}
\end{algorithm}

The choice of the 50 computed components is arbitrary, as 10
components incorporate $99.9\%$ of the Frobenius norm of the
trajectory matrix $\X$.  The RMSE of the linear regression model in
the ESPRIT indicates the accuracy of the model~(\ref{eq:1}). For a set
of 50 EL images, the mean RMSE equals 0.033.

\begin{figure*}
  \centering
  \subfloat[Original image\label{fig:comp original}]{%
    \includegraphics[width=.4\textwidth]{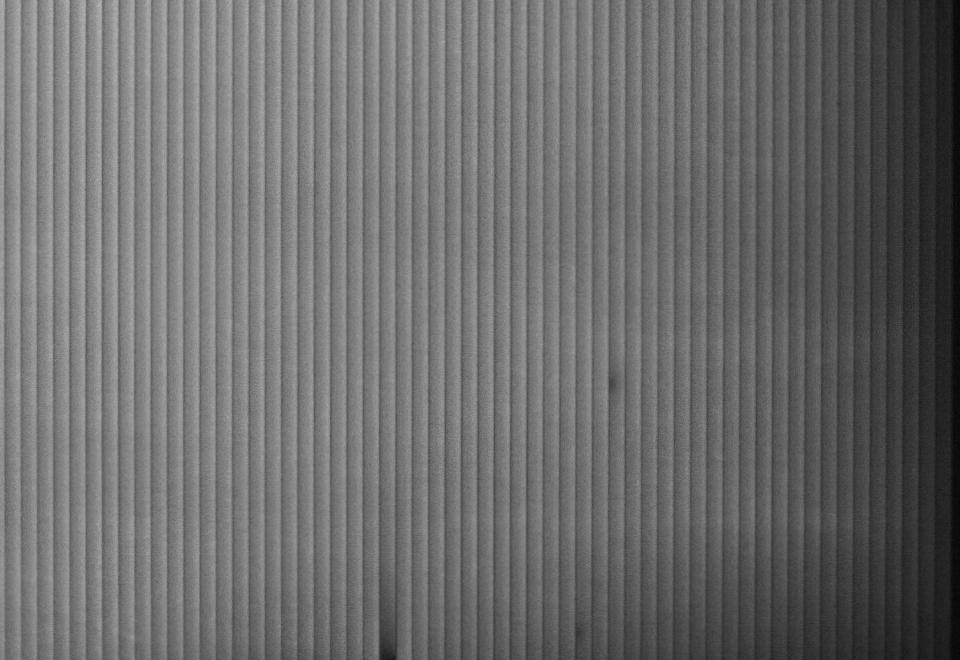}
  }
  \hspace{12pt}
  \subfloat[Cell component\label{fig:comp cell}]{%
    \includegraphics[width=.4\textwidth]{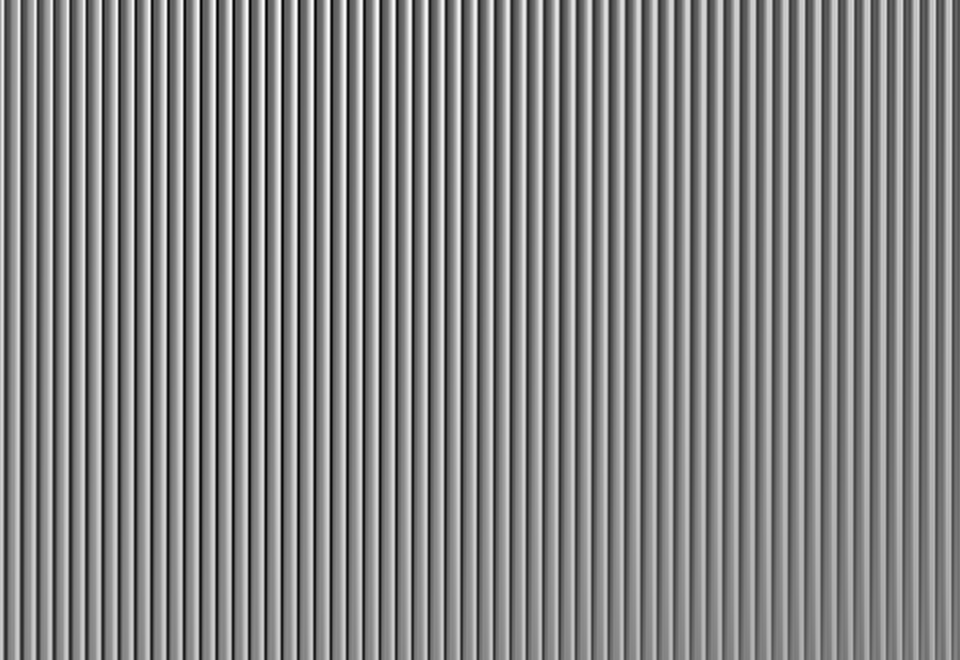}
  }
  \hfill
  \subfloat[Global variations component\label{fig:comp noncell}]{%
    \includegraphics[width=.4\textwidth]{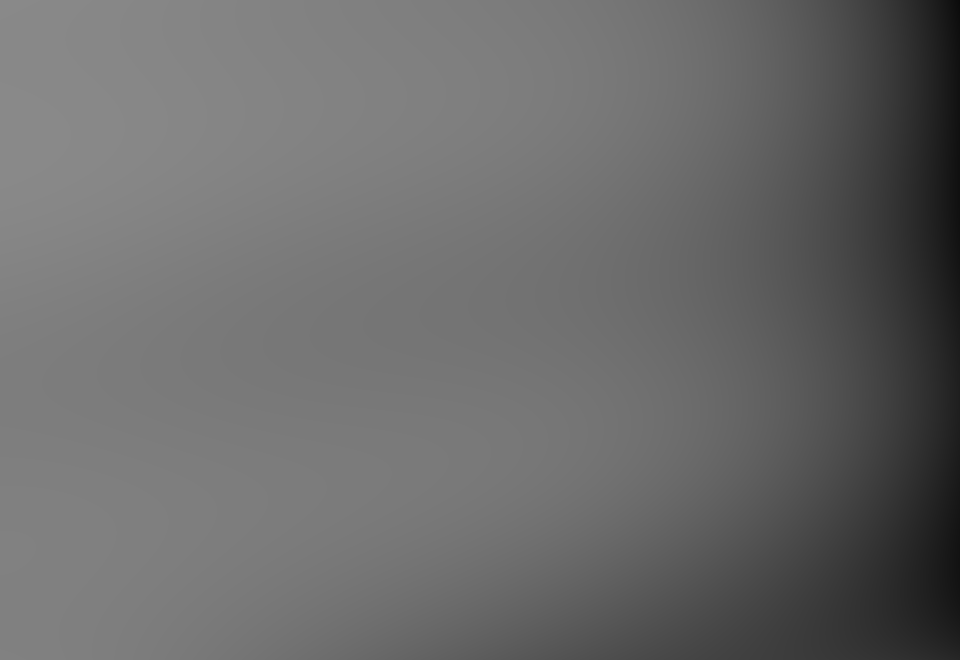}
  }
  \hspace{12pt}
  \subfloat[Aperiodic component\label{fig:comp residual}]{%
    \includegraphics[width=.4\textwidth]{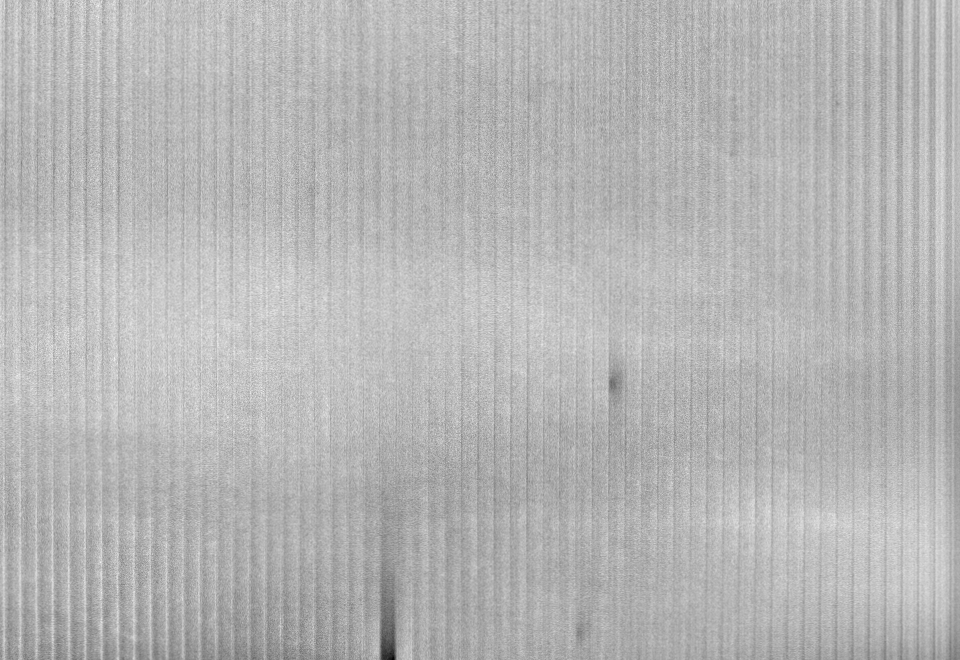}
  }
  \caption{Decomposition of an EL image onto 3 components}
  \label{fig: ssa output}
\end{figure*}

Figure~\ref{fig: ssa output} depicts the non-cell components, the cell
components, and the residual image. It can be argued that different
components have a different physical origin. By its nature, the global
variation component describes changes in a material which results in
large losses that spread over large portions of a module. The cell
component variation is influenced by the EL measurement
conditions. Lastly, the aperiodic component captures effects caused by
non-regular changes in material like shunts, or droplets (see
\cite{sovetkin2020encoder}).\linelabel{physical interpretation of components}


Shunts are characterized by a more conductive connection between the
front and back electrodes than the normal solar cell structure (i.e.\
the solar cell structure is damaged or missing). There are many causes
for shunts. Commonly shunts originate from debris of the copper
evaporation source or pinholes in the CIGS absorber
\cite{misic2015debris, misic2017thesisoriginofshunts}. 
Shunts are generally relevant to the solar module performance, in
particular under low light conditions
\cite{weber2011electroluminescence}.

In addition to shunts we noticed the CIGS modules often exhibit
``droplets'' in the EL images.  The appearance of droplets resembles
water stains and thus we speculate these structures originate from the
chemical bath deposition. At this point it is unknown what the impact
of droplets is on the module performance, however, the bright
appearance imply a local change in quantum efficiency according to the
reciprocity relations between luminescence and quantum efficiency
\cite{rau2007reciprocity}.

We remark that decomposition can be applied to an image (additive
model assumption\linelabel{additive and multiplicative models}) as
well as to the logarithm of an image (multiplicative model
assumption). The logarithm of an image corresponds to the internal
voltage (see~\eqref{eq:4}, Section~\ref{sec:char-length-estim}).

Furthermore, we remark that it is important to correct any perspective
distortion present in images, as a periodic image distorted in such a
way is no longer a signal with a finite rank in the settings of the
2D-SSA methodology.

\subsection{Interconnection line detection}
\label{sec:laser-line-detection}

In order to identify interconnection lines, it is sufficient to locate
the global minima for every level in the normal direction of the
interconnection lines.

The ESPRIT model satisfies equation~(\ref{eq:1}), which is a sum of
amplitude-modulated cosine functions. A derivative of such function is
again a sum of polynomial amplitude-modulated cosine functions,
similar to the general form of the signal of finite
rank~(\ref{eq:2}). Such derivatives can be computed
symbolically. Hence all the global minima in the normal direction of
the interconnection lines can be identified by evaluating precise
values of the derivatives, and filtering out points that satisfy a
minima extreme point requirement.

Algorithm~\ref{algo:laser line detection} describes the steps of the
interconnection line identification. Symbolic derivatives are computed
using the ``Deriv'' R-package, \cite{derivR2019}.

\begin{algorithm}
  {\bf Input:}
  \begin{itemize}
  \item EL image with dimensions $N_{x} \times N_{y}$.
  \item Regular mesh of points
    $P = \{n,m \in \mathbb{R}, 1 \leq n \leq N_{x}, 1 \leq m \leq
    N_{y}\}$.
  \end{itemize}

  {\bf Output:} Array of coordinates $O \subset P$, corresponding to the
  estimated locations of interconnection lines.

  \begin{enumerate}
  \item Compute EL image decomposition with
    Algorithm~\ref{algo:decomposition}.
  \item Use ESPRIT and linear regression to estimate parameters of the
    model~(\ref{eq:1}), see Sections~\ref{sec:esprit} and
    \ref{sec:ampl-phase-estim}. Denote the resulting signal as
    $S: P \to \mathbb{R}$.
  \item Compute symbolically expression for derivative
    $dS \coloneqq \frac{\partial S}{\partial m}: P \to \mathbb{R}$.
  \item The output set
    $O \coloneqq \{ p \in P : dS(p-) < 0 \text{ and } dS(p+) > 0\}$,
    where $p-, p+$ are the neighbours in $P$ of the point $p$ in the
    $m$-variable direction.
  \end{enumerate}
  \caption{Interconnection line detection\label{algo:laser line
      detection}}
\end{algorithm}

We remark that the resulting expression for the cell component signal,
as well as its derivatives, can be computed on a finer grid $P$ than
the original pixel coordinates. Hence the interconnection line
identification is performed with sub-pixel accuracy.

\begin{figure*}
  \centering \subfloat[Isolated cell components\label{fig: cell
    component}]{%
    \includegraphics[width=.3\linewidth]{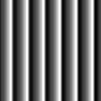}}
  \hspace{12pt} \subfloat[Pixel value intensity in the normal
  direction to the interconnection lines\label{fig: local minima}]{%
    \includegraphics[width=.5\linewidth]{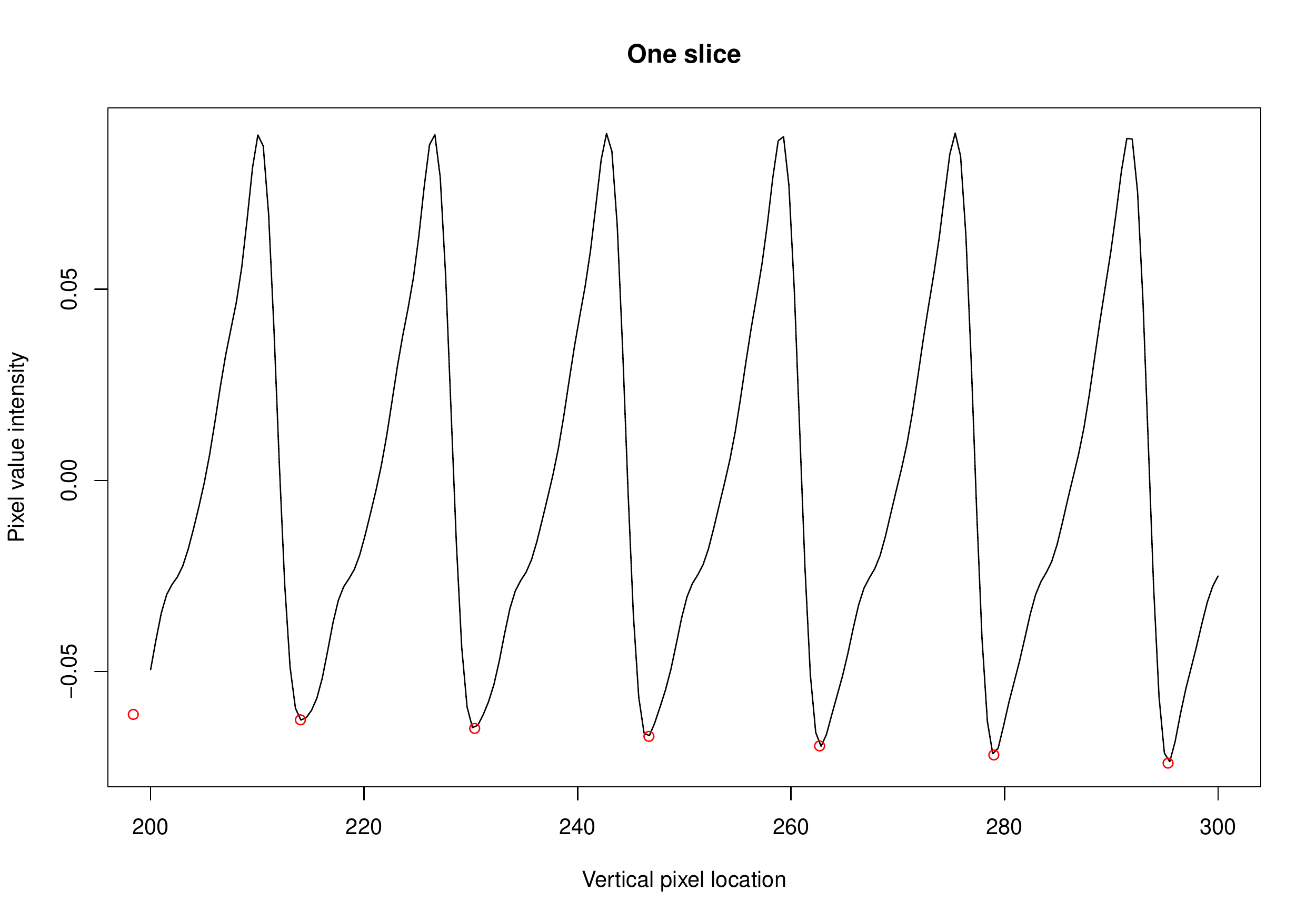}}
  \hfill \subfloat[A cell with indicated interconnection lines
  (red)\label{fig: lines shown}]{%
    \includegraphics[angle=270,
    width=0.65\linewidth]{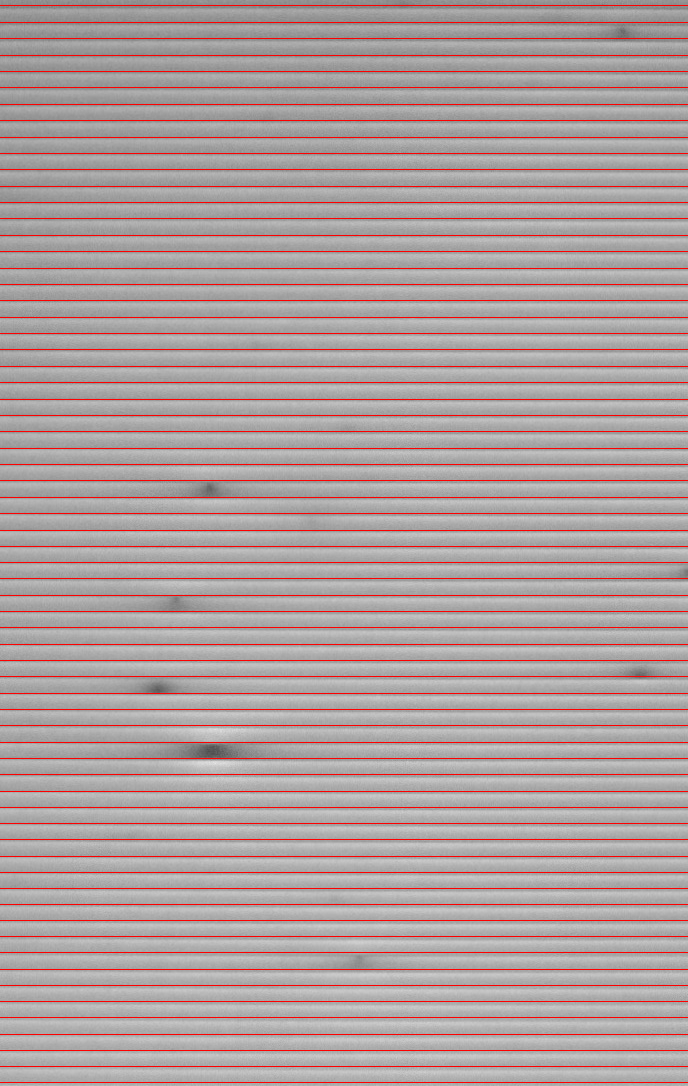}}
  \caption{Detection of the interconnection lines\label{fig: laser
      line}}
\end{figure*}

Figure~\ref{fig: laser line} visually demonstrates the steps of the
method. Figure~\ref{fig: cell component} displays the estimated cell
component. Figure~\ref{fig: local minima} shows a slice of a pixel
value intensities in Figure~\ref{fig: cell component}, where computed
local minima identified with red dots. Lastly, Figure~\ref{fig: lines
  shown} shows the module with its interconnection lines identified by
red lines.

\subsection{Inverse characteristic length estimation}
\label{sec:char-length-estim}

Another application of the obtained parametric expression for a module
is an estimation of the inverse characteristic length $\lambda$
\cite{helbig2010quantitative,pieters2015cellintensity}.

In a defect-free PV module, there are no currents in the vertical
direction (along the interconnection line direction), and thus current
satisfies the following linearised 1D Poisson equation,
\cite{helbig2010quantitative,augarten2016calculation}.
\begin{equation}
  \label{eq:3}
  \frac{\partial^{2} V}{\partial x^{2}} = \lambda^{2} V,
\end{equation}
where $\lambda = \sqrt{\frac{R_{\mathrm{sheet}}}{r_\textrm{j}}}$ is
the inverse characteristic length, where $R_{\mathrm{sheet}}$ is the
sheet resistance and $r_\textrm{j}$ is the local differential junction
resistance. Note, that (\ref{eq:3}) is an equation between two fields,
where $V(x,y)$ and $\lambda(x,y)$ depend on the position $(x,y)$ in a
PV module.

The luminescence intensity related\linelabel{intensity and voltage} to the internal voltage via the
following relation
\begin{equation}
  \label{eq:4}
  I = c \exp(Vc_{0}) \implies V = \frac{1}{c_{0}} \ln(I/c),
\end{equation}
where the constant $c_{0} = \frac{q}{kT}$ is the thermal voltage (with
the elementary charge, $q$, Boltzmann's constant, $k$, and the
temperature, $T$), $c$ is a parameter that describes the optical
system from the photon generation in the solar cell absorber material
to the photon detection within the camera. As such the parameter $c$
depends on the quantum efficiency of both the solar cell and the used
camera including all optical components, and the spectral photon
density of a black body \cite{rau2007reciprocity}. Generally, the
constant $c$ may vary over the solar cell area due to the camera
optics and variations in the solar cell properties. However, in
general, the variations in $c$ are small compared to the exponential
voltage-dependency \cite{abou2016advanced}. In our analysis, we assume
that $c$ is constant over the whole module area, and is accessible to
a researcher.

The decomposition of a signal onto trend and cell components provides
us an EL image signal without small aperiodic defects like shunts
(small dark areas within cells boundaries). The global intensity and
cell components without small aperiodic defects can be considered as a
module without defects. Hence, we consider a module without defects to
be an image $G + S$, of the output of
Algorithm~\ref{algo:decomposition}.

Hence combining (\ref{eq:3}) and (\ref{eq:4}), allows us to express
inverse characteristic length as a function of intensity and its
derivative:
\begin{equation}
  \frac{\partial^{2} V}{\partial x^{2}} =
  \frac{
    \frac{\partial^{2} I}{\partial x^{2}} I - \left(\frac{\partial I}{\partial x}\right)^{2}
  }{
    c_{0} I^{2},
  }
\end{equation}
and therefore,
\begin{equation}
  \lambda^{2} =
  \frac{
    \frac{\partial^{2} I}{\partial x^{2}} I - \left(\frac{\partial I}{\partial x}\right)^{2}
  }{
    I^{2} \ln(I/c)
  }.
\end{equation}

All derivatives are computed symbolically from the estimated cell
component signal.

\subsection{Stitched image correction}
\label{sec:stitch-image-corr}

\begin{figure*}
  \centering
  \subfloat[Original stitched image\label{fig: stitched}]{%
    \includegraphics[width=.35\linewidth]{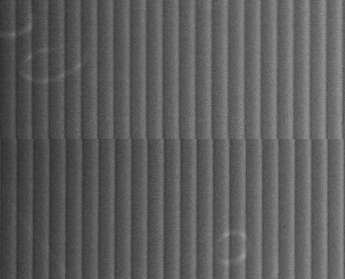}}
  \hspace{12pt}
  \subfloat[Corrected image\label{fig: stitched corrected}]{%
    \includegraphics[width=.35\linewidth]{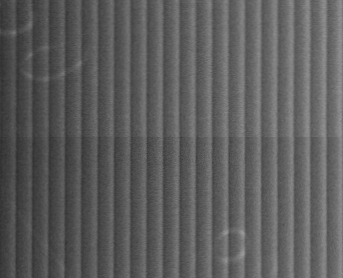}}
  \caption{An example of the stitched image correction\label{fig:
      stitch algo}}
\end{figure*}

In order to achieve higher image resolution, several EL images can be
stitched together. However, the image aligning can be imperfect, as
shown in Figure~\ref{fig: stitched}. These misaligned image patches
and the resulting stitch line can be attributed to incorrect
perspective as well as radial distortions.  The latter distortion
leads to a non-linear transformation that is needed to be applied to
an image for the stitch line correction.

Algorithm~\ref{algo:stitched image} proposes an approach to correct
such distortion. The basic idea of the algorithm is the estimation of
phase shifts in neighbouring locations, where each shift is estimated
by application of MSSA, \cite{golyandina2018singular}, in a direction
perpendicular to the interconnection lines.

\begin{algorithm}
  {\bf Input:} EL image $X$ with dimensions $N_{x} \times N_{y}$.

  {\bf Output:} A displacement map $M$, a vector of dimension
  $N_{x}-1$ indicating size of horizontal shifts in image $X$ (except
  for the first row).

  \begin{enumerate}
  \item For each image row $i$ starting from the second row:
    \begin{enumerate}
    \item Compute MSSA of the $i$ and $i-1$ row of the image $I$.
    \item Estimate the ESPRIT parameters.
    \item Filter out periods not corresponding to cell components. Let
      cell component indices be $I$.
    \item Compute shift for cell each component
      $s_{k} \coloneqq \frac{\phi_{k}}{2\pi \omega_{k}}, k \in I$.
    \end{enumerate}
  \item Set $M_{i} \coloneqq \max_{k \in I} s_{k}$.
  \end{enumerate}
  \caption{Stitched image correction\label{algo:stitched image}}
\end{algorithm}

The result of the algorithm is a displacement map that defines shifts
in the horizontal direction. Note that shifts have sub-pixel accuracy.

Figure~\ref{fig: stitch algo} shows the result of the
algorithm. Figure~\ref{fig: stitched} shows a patch of the original EL
image with a stitched part. Figure~\ref{fig: stitched corrected}
depicts the shifted image, resulting from applying the displacement
map. As each displacement is estimated locally between neighbours, the
resulting transformation is a non-linear transformation.

To evaluate the accuracy of the phase shift estimation we model the
following two time series
\begin{align}
  s_{1}(x) = \cos(2 \pi x/50) & + \cos(2 \pi x/20) + \cos(2 \pi x/30) & + \varepsilon_{1}(x), \\
  s_{2}(x) = 2\cos(2 \pi x/70) & + \underbrace{\cos(2 \pi (x + 7)/20) + \cos(2 \pi (x + 7)/30)}_{\text{signal}} & + \varepsilon_{2}(x), \\
\end{align}
where $x \in \{1,2,\ldots,1000\}$, periodics with period $20$ and $30$
correspond to the signal (models the cell component), periodics with
period $50$ and $70$ slowly varying trend, and
$\varepsilon_{1}, \varepsilon_{2}$ are two independent Gaussian iid
processes with zero-mean and unit variance.

\begin{figure}
  \sidecaption
  \centering
  \includegraphics[width=0.6\textwidth]{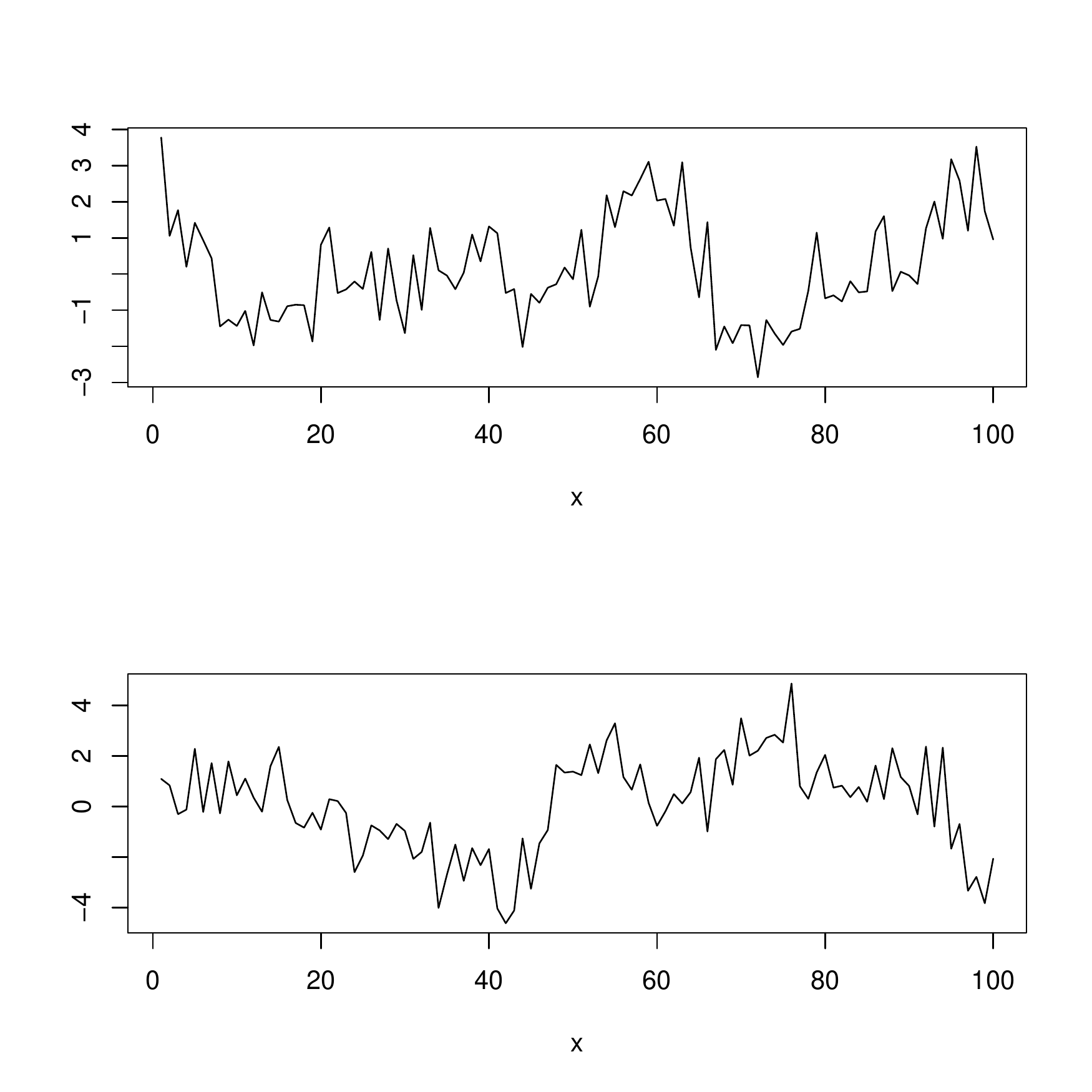}
  \caption{$s_1(x)$ and $s_2(x)$ time series, for $x \in [1,100]$. The
    signal (periodics with periods 20 and 30) are ``hidden'' with a
    trend and a noise components\label{fig: s1 and s2}}
\end{figure}

The signal part of the series $s_{1}$ is shifted by $7$ units relative
to the signal of the series $s_{2}$. Figure~\ref{fig: s1 and s2} shows
a part of those series on the interval $[1,100]$.

Table~\ref{table: stitch accuracy} shows the accuracy of the shift
estimation of the selected signal components. That simulation was
performed using 100 repetitions.

\begin{table}
  \centering
  \caption{Accuracy of the shift estimation between signals in time
    series $s_{1}$ and $s_{2}$\label{table: stitch accuracy}}
  \begin{tabular}{c|c|c|c}
    RMSE & Estimate mean & 25\% quantile & 75\% quantile\\
    \hline
    0.248 & 7.003 & 6.836 & 7.160
  \end{tabular}
\end{table}

\section{Conclusions}
\label{sec:conclusions}

In this paper we demonstrated an application of SSA on EL images of
thin-film PV modules. This low-rank approach allows capturing several
important aspects of those images, namely, global and local repetitive
variations (or cell components) in an EL image.

Several image processing algorithms based on parametric models of SSA
are proposed. The first method identified the interconnection lines
between the individual cells at sub-pixel accuracy, and the second
method corrects the incorrectly stitched images.

Furthermore, we propose an approach based on symbolic differentiation
of the SSA signal to estimate the so-called inverse characteristic
length, a physical parameter of a module.

We note that in the settings of 2D-SSA it is important to correct
perspective distortion in such EL images. The two-dimensional cosine
array transformed with perspective distortion is no longer a
finite-rank signal.

It should be noted that the information captured by SSA is not
complete, as local aperiodic features, such as shunts and droplets are
not signals of finite rank. Therefore, the full analysis of such PV
modules requires other methods. For example, we complement the signal
captured by SSA using an encoder-decoder segmentation approach of
individual defects (see
\cite{sovetkin2020encoder,sovetkin2020pvaided}).

Lastly, we remark that source code and a sample of EL images data are
available upon request.

\begin{acknowledgement}
This work is supported by the Solar-era.net framework in the project
``PEARL TF-PV'' (Förderkennzeichen: 0324193A) and partly funded by the
HGF project ``Living Lab Energy Campus (LLEC)''.
\end{acknowledgement}

\section*{Appendix}
\addcontentsline{toc}{section}{Appendix}

\subsection*{2D-SSA embedding and projection}

Following notations of \cite{golyandina20102d}, let $x$ be an 2D-array
with dimensions $(N_{x},N_{y}) \in \mathbb{N}^{2}$:
\begin{equation}
  \footnotesize
  x = \left(
    \begin{array}{cccc}
      x(0,0) & x(0,1) & \ldots & x(0,N_{y-1}) \\
      x(1,0) & x(1,1) & \ldots & x(1,N_{y-1}) \\
      \vdots & \vdots & \ddots & \vdots \\
      x(N_{x-1},0) & x(N_{x-1},1) & \ldots & x(N_{x-1},N_{y-1}) \\
    \end{array}
  \right).
\end{equation}

The 2D-SSA embedding is defined by the window size vector
$(L_{x},L_{y})$, which is restricted by
$1 \leq L_{x} \leq N_{x}, 1\leq L_{y} \leq N_{y}$, and
$1 < L_{x}L_{y} < N_{x}N_{y}$. Let $K_{x} \coloneqq N_{x} - L_{x} +1$
and $K_{y} \coloneqq N_{y} - L_{y} + 1$, then the trajectory matrix of
2D-SSA is given by the following Hankel-block-Hankel matrix:
\begin{equation}
  \footnotesize
  \X \coloneqq \left(
    \begin{array}{ccccc}
      \X_0 & \X_1 & \X_2 & \ldots &\X_{K_y - 1} \\
      \X_1 & \X_2 & \X_3 & \ldots &\X_{K_y} \\
      \X_2 & \X_3 & \ddots & \ddots  & \vdots \\
      \vdots & \vdots & \ddots & \ddots & \vdots \\
      \X_{L_y - 1} & \X_{L_y} & \ldots & \ldots  & \X_{N_y - 1} \\
    \end{array}
  \right),
\end{equation}
where each block
\begin{equation}
  \footnotesize
  \X_i \coloneqq \left(
    \begin{array}{cccc}
      x(0,i) & x(1,i) & \ldots & x(K_x - 1, i) \\
      x(1,i) & x(2,i) & \ldots & x(K_x, i) \\
      \vdots & \vdots & \ddots & \vdots \\
      x(L_x - 1,i) & x(L_x,i) & \ldots & x(N_x - 1, i) \\
    \end{array}
  \right).
\end{equation}

By construction there is a one-to-one correspondence between 2D-arrays
of size $N_{x} \times N_{y}$ and the Hankel-block-Hankel matrices.

Let $\Z$ be an arbitrary matrix with a block-structure, where each
block $\Z_{i}$ has the same dimension as the matrix $\X_{i}$
\begin{equation}
  \footnotesize
  \mathbf{Z} = \left(
    \begin{array}{ccccc}
      \Z_0 & \Z_1 & \Z_2 & \ldots &\Z_{K_y - 1} \\
      \Z_1 & \Z_2 & \Z_3 & \ldots &\Z_{K_y} \\
      \Z_2 & \Z_3 & \ddots & \ddots  & \vdots \\
      \vdots & \vdots & \ddots & \ddots & \vdots \\
      \Z_{L_y - 1} & \Z_{L_y} & \ldots & \ldots  & \Z_{N_y - 1} \\
    \end{array}
  \right).
\end{equation}
Then a projection of $\Z$ to Hankel-block-Hankel matrix can be
computed in two steps. Firstly, diagonal averaging is performed within
blocks $\Z_{i}, i \in 0,\ldots, N_{y} - 1$. Secondly, the blocks of
the matrix $Z$ are averaged between themselves. The projection can be
applied in the reverse order as well.

\subsection*{SVD}

Let $\mathbf{S} = \X \X^{T}$,
$\lambda_{1} \geq \ldots \geq \lambda_{d} > 0$ be non-zero eigenvalues
of the matrix $\mathbf{S}$, $U_{1},\ldots,U_{d}$ be the corresponding
eigenvectors, and
$V_{i} \coloneqq \X^{T} U_{i} / \sqrt{\lambda_{i}}, i = 1,\ldots,d$.

Then SVD of the matrix $\X$ can be written as
\begin{equation}
  \X = \X_1 + \ldots + \X_d.
\end{equation}
The values $\sqrt{\lambda_{i}}$ are the singular values of $\X$.

\subsection*{Finite-rank signal}

An image $x$ has rank $r$ if the rank of trajectory matrix $\X$ equals
$r < \min(L_{x},L_{y},K_{y},K_{y})$. In other words, the trajectory
matrix is rank-deficient. If rank $r$ does not depend on the choice of
$L$ for any sufficiently large dimensions of $x$, then $x$ is called
to have a {\it finite rank}.

Objects of finite rank are closely related to the linear recurrent
sequences, \cite{kurakin1995linear}. Linear recurrent formulae are
used to build forecast based on the SSA signal.

\subsection*{Computational complexity}
\linelabel{computational complexity}

From the computational point of view, the hardest steps of the
proposed algorithms are the singular value decomposition (SVD), ESPRIT
(Section~\ref{sec:esprit}), and the phase and amplitude least-squares
fit (Section~\ref{sec:ampl-phase-estim}).

In the context of the SSA application, the computational complexity of
the SVD of a Hankel-block-Hankel matrix is $O(k N \log N + k^{2} N)$,
where $N$ is the number of pixels in an image and $k$ is the number of
computed eigentriples, \cite{korobeynikov2010computation}. The linear
equations solved for ESPRIT and the phase and amplitude parameters
estimations are solved using QR-decomposition, and require $O(N^{3})$
operations.  Our numerical experiments show that the SVD decomposition
dominates the computational time for the proposed algorithms for image
sizes with width and height less than 4000 pixels.

To measure the required time and memory of the proposed algorithms we
utilise a machine with Intel(R) Xeon(R) CPU E5-1620 3.5GHz processor
and 31GB of RAM. The time measurements are performed on a single CPU.

\begin{figure}
  \sidecaption
  \centering
  \includegraphics[width=0.8\textwidth]{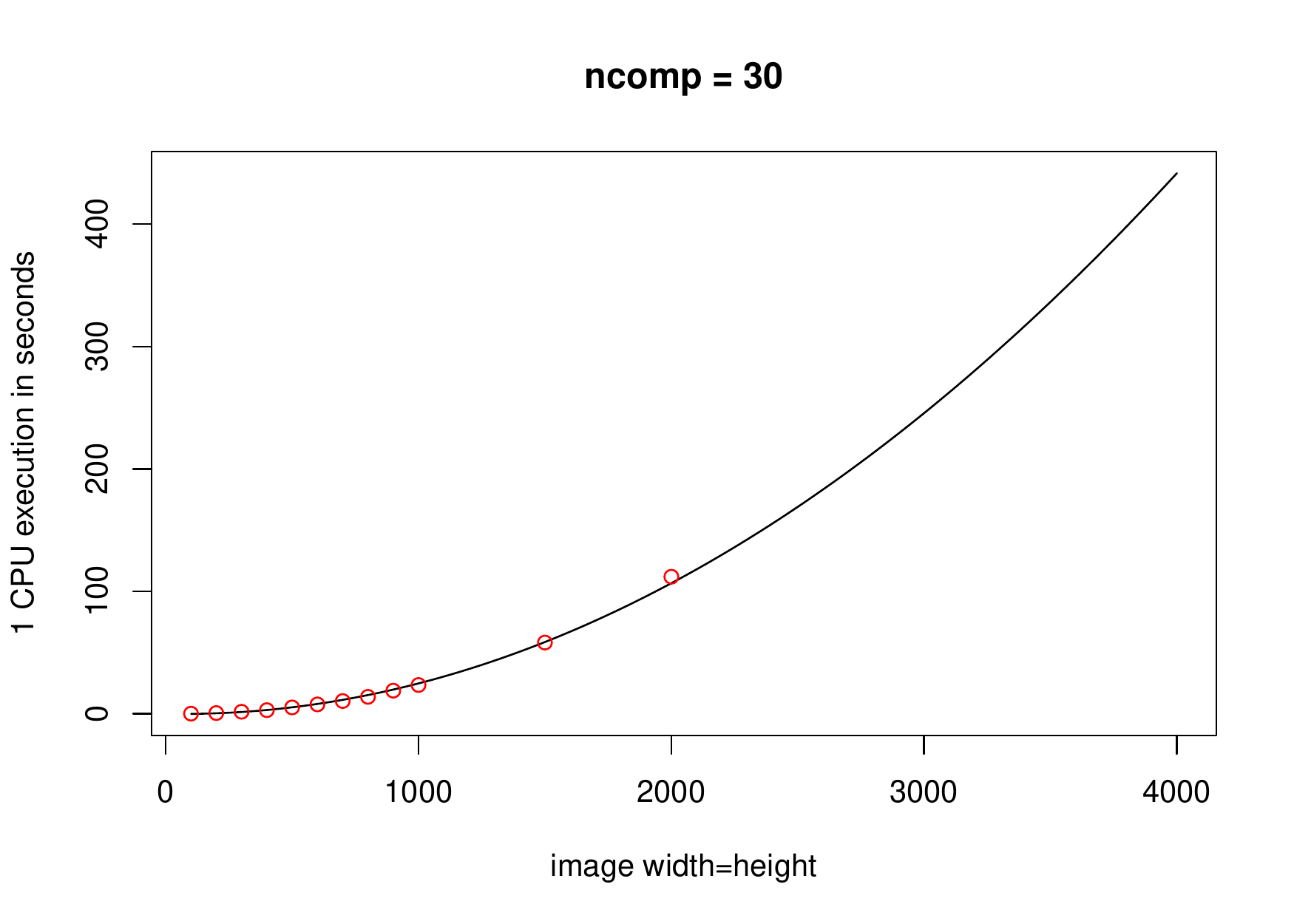}
  \caption{Computation time required for
    Algorithm~\ref{algo:decomposition} for different input image
    sizes.\label{fig: execution time}}
\end{figure}

Figure~\ref{fig: execution time} shows the time required to perform
steps of Algorithm~\ref{algo:decomposition} for square images (width
equals height) for different image widths. The red points correspond
to the measured time in seconds, and the black line is a parabola
fitted to the points. Table~\ref{fig: ram memory} shows the amount of
memory required for the steps of Algorithm~\ref{algo:decomposition}.

\begin{table}
  \centering
  \caption{Maximum memory required for
    Algorithm~\ref{algo:decomposition} for different input image
    sizes.\label{fig: ram memory}}
  \begin{tabular}{c|c|c|c|c|}
    Image width and height & 500 & 1000 & 1500 & 2000 \\
    \hline
    Maximum RAM in GB & 0.5 & 1.6 & 3.2 & 5.5 \\
  \end{tabular}
\end{table}

Algorithm~\ref{algo:laser line detection} utilises results of the
Algorithm~\ref{algo:decomposition} and requires additional symbolical
differentiation. Such computation is also required for the inverse
characteristic length (Section~\ref{sec:char-length-estim}). The
running time of the symbolical differentiation depends on the number
of terms of the signal~(\ref{eq:1}). Table~\ref{table: running time
  diff} demonstrates the relationship between the number of terms in
the signal and the running time required for the symbolic
differentiation. Algorithm~\ref{algo:laser line detection} and the
inverse characteristic length computation works with the cell
component that typically consists of 5--7 terms.

\begin{table}
  \centering
  \caption{Running time of the symbolical differentiation routine used
    in Algorithm~\ref{algo:laser line detection} and the inverse
    characteristic length estimation in Section~\ref{sec:char-length-estim}\label{table: running time diff}}
  \begin{tabular}{c|c|c|c|c|c}
    Number of components & 5 & 9 & 13 & 17 \\
    \hline
    Execution time in seconds & 0.15 & 0.46 & 1.27 & 1.79 \\
  \end{tabular}
\end{table}

Algorithm~\ref{algo:stitched image} utilises a different version of
SSA, that requires less time and memory, however, the algorithm
computes MSSA multiple times for several pairs of neighbour rows. A
single iteration of the algorithm loop requires 0.7 seconds and 0.78
of RAM for an image with width of 4000 pixels. Different iterations of
the loop can be run in parallel. Note that most of the used memory is
occupied by an image itself, and hence the memory can be shared by
multiple processes.  Algorithm~\ref{algo:stitched image} usually
requires running about 100--200 iterations, as an approximate location
of the stitching lines is known. Hence, the total running time an
8-core processor can be as little as 20 seconds.

Lastly, we remark the proposed algorithms run in a deterministic
amount of time. Hence, the methods can be run in real-time
applications.\linelabel{real time remark}

\bibliographystyle{spbasic}
\bibliography{bib.bib}

\end{document}